\documentclass{article}

 \usepackage[preprint]{neurips_2026}

\usepackage[utf8]{inputenc} % allow utf-8 input
\usepackage[T1]{fontenc}    % use 8-bit T1 fonts
\usepackage{hyperref}       % hyperlinks
\usepackage{url}            % simple URL typesetting
\usepackage{booktabs}       % professional-quality tables
\usepackage{amsfonts}       % blackboard math symbols
\usepackage{nicefrac}       % compact symbols for 1/2, etc.
\usepackage{microtype}      % microtypography
\usepackage{xcolor}         % colors

\usepackage{amsmath}
\usepackage{amssymb}
\usepackage{mathtools}
\usepackage{amsthm}

\usepackage{multirow}    % 提供 \multirow
\usepackage[table]{xcolor} % 提供 \cellcolor 和 颜色支持
\usepackage[dvipsnames, table]{xcolor}
\usepackage{wrapfig}
\usepackage[capitalize,noabbrev]{cleveref}
\usepackage{float}
\usepackage{placeins}
\usepackage{bbding}
\title{EvoDriveVLA: Evolving Driving VLA Models via Collaborative Perception-Planning Distillation}

\author{
  \textbf{Jiajun Cao}$^{1,2^*}$ \quad \textbf{Xiaoan Zhang}$^{1,2^*}$ \quad \textbf{Xiaobao Wei}$^{1^*}$ \quad \textbf{Liyuqiu Huang}$^{1,2}$ \quad \textbf{Zijian Wang}$^{2}$ \\
  \textbf{Hanzhen Zhang}$^{2}$ \quad \textbf{Zhengyu Jia}$^{2}$ \quad \textbf{Wei Mao}$^{2}$ \quad \textbf{Hao Wang}$^{1}$ \quad \textbf{Xianming Liu}$^{2}$ \\
  \textbf{Shuchang Zhou}$^{2}$ \quad \textbf{Yang Wang}\textsuperscript{2\Envelope} \quad \textbf{Shanghang Zhang}\textsuperscript{1\Envelope} \\[0.5em]
  $^1$State Key Laboratory of Multimedia Information Processing, School of Computer Science, \\
  Peking University \quad
  $^2$XPeng Motors \\
}

\begin{document}

\maketitle

{\let\thefootnote\relax\footnotetext{$^{*}$ Equal contribution, \textsuperscript{\Envelope} Corresponding author}}

\begin{abstract}
    Vision-Language-Action models have shown great promise for autonomous driving, yet they suffer from degraded perception after unfreezing the visual encoder and struggle with accumulated instability in long-term planning. To address these challenges, we propose \textbf{EvoDriveVLA}—a novel collaborative perception-planning distillation framework that integrates self-anchored perceptual constraints and future-informed trajectory optimization. Specifically, self-anchored visual distillation leverages self-anchor teacher to deliver visual anchoring constraints, regularizing student representations via trajectory-guided key-region awareness. In parallel, future-informed trajectory distillation employs a future-aware oracle teacher with coarse-to-fine trajectory refinement and Monte Carlo dropout sampling to synthesize reasoning trajectories that model future evolutions, enabling the student model to internalize the future-aware insights of the teacher. EvoDriveVLA achieves SOTA performance in nuScenes open-loop evaluation and significantly enhances performance in  NAVSIM closed-loop evaluation. 
    % Our code will be released.
    Our code is available at:
    \href{https://github.com/hey-cjj/EvoDriveVLA}{https://github.com/hey-cjj/EvoDriveVLA}.
\end{abstract}

\section{Introduction}

\begin{figure*}[t]
    \centering
    \includegraphics[width=0.9\textwidth]{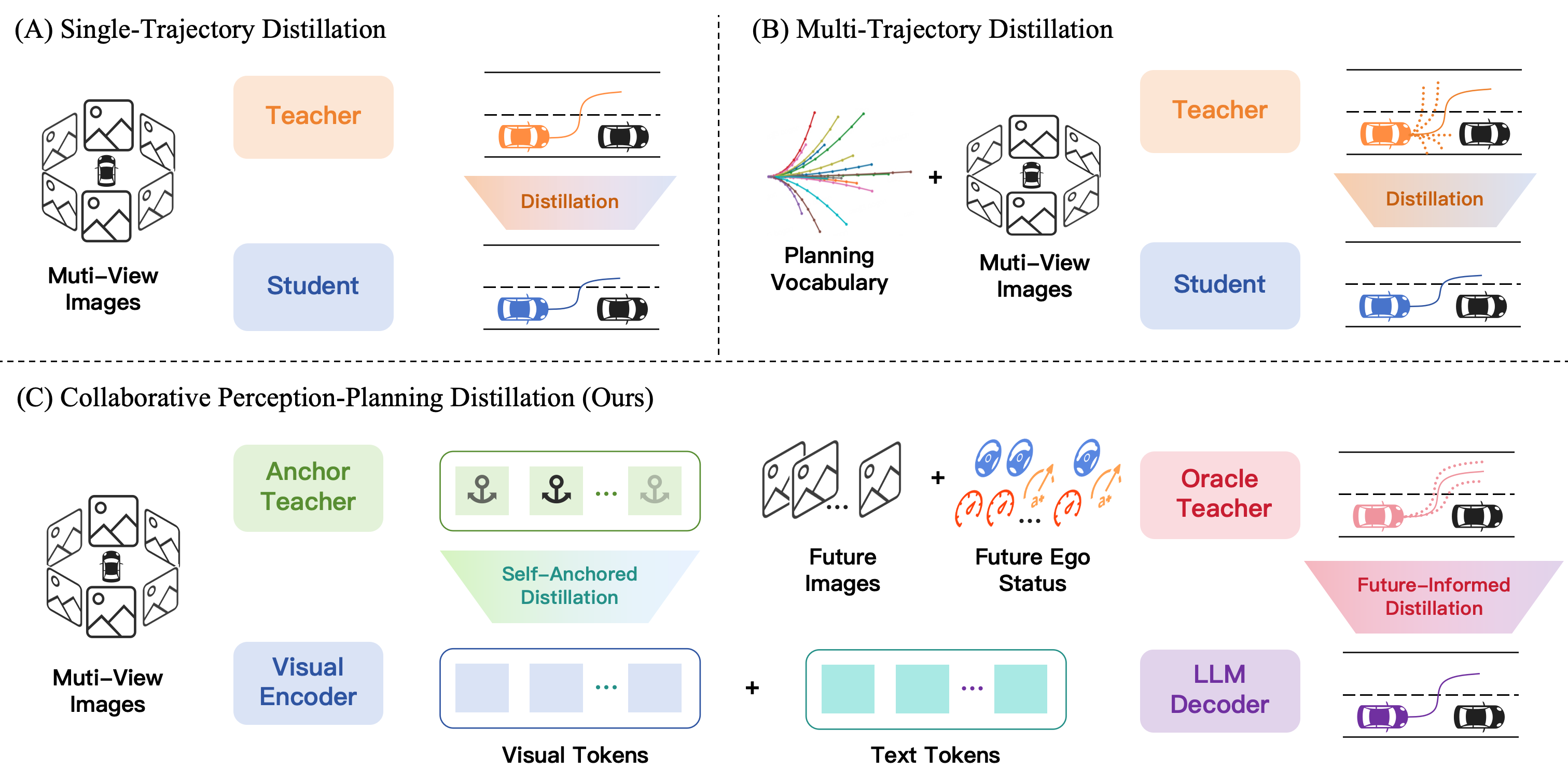}
    \vspace{-0mm}
    \caption{Comparison of distillation methods for autonomous driving. (a) Single-Trajectory Distillation; (b) Multi-Trajectory Distillation; (c) Collaborative Perception-Planning Distillation (Ours).}
    \label{fig:compare}
    \vspace{-3mm}
\end{figure*}

With the rapid advances of Vision-Language Models (VLMs)~\cite{liu2023visual, bai2025qwen2, zhang2025beyond}, increasing attention has been directed towards leveraging VLMs for autonomous driving, giving rise to driving Vision-Language-Action (VLA) models that can directly output driving actions and trajectories. Compared to traditional end-to-end approaches~\cite{hu2023planning, jiang2023vad, gao2025rad}, VLA models are capable of not only predicting trajectories, but also understanding navigation instructions~\cite{peng2025navigscene}, performing scene-based question answering~\cite{sima2024drivelm}, and utilizing chain-of-thought reasoning~\cite{li2025recogdrive}. Their superior generalization and reasoning potential make them the mainstay in autonomous driving. However, during practical training, VLA models suffer from degraded perceptual capabilities after unfreezing the visual encoder, as well as trajectory instability in long-term planning.

As a pivotal technique for boosting the performance of autonomous driving systems, knowledge distillation~\cite{hinton2015distilling} has gained significant traction in recent research. As illustrated in Fig.~\ref{fig:compare}, existing distillation methods can be categorized into single-trajectory distillation and multi-trajectory distillation. Single-trajectory approaches, exemplified by DiMA~\cite{hegde2025distilling}, directly supervise the student using trajectories predicted by a teacher model. In contrast, multi-trajectory methods, such as DistillDrive~\cite{yu2025distilldrive}, encourage the teacher to produce diverse trajectory outputs by constructing a planning vocabulary, aiming to enrich planning knowledge in distillation through structured trajectory candidates and alleviate the limited expressiveness and poor scenario adaptability of single-trajectory approaches.

However, existing methods have not considered sufficiently the principled design of knowledge distillation for autonomous driving:
(1) The visual encoder, which serves as the core component of scene perception, has not been adequately emphasized or effectively handled during the distillation process in existing training pipelines.
(2) When the teacher and student models are trained under identical settings, the teacher offers no substantial advantage in planning capability, and therefore fails to provide more accurate or informative knowledge for distillation.
(3) Although existing multi-trajectory distillation methods increase the diversity of teacher-generated trajectories, such diversity is largely constrained by predefined planning vocabularies, limiting their ability to truly adapt to the dynamic and context-dependent nature of real-world driving scenarios.

\begin{figure*}[t]
    \centering
    \includegraphics[width=0.99\textwidth]{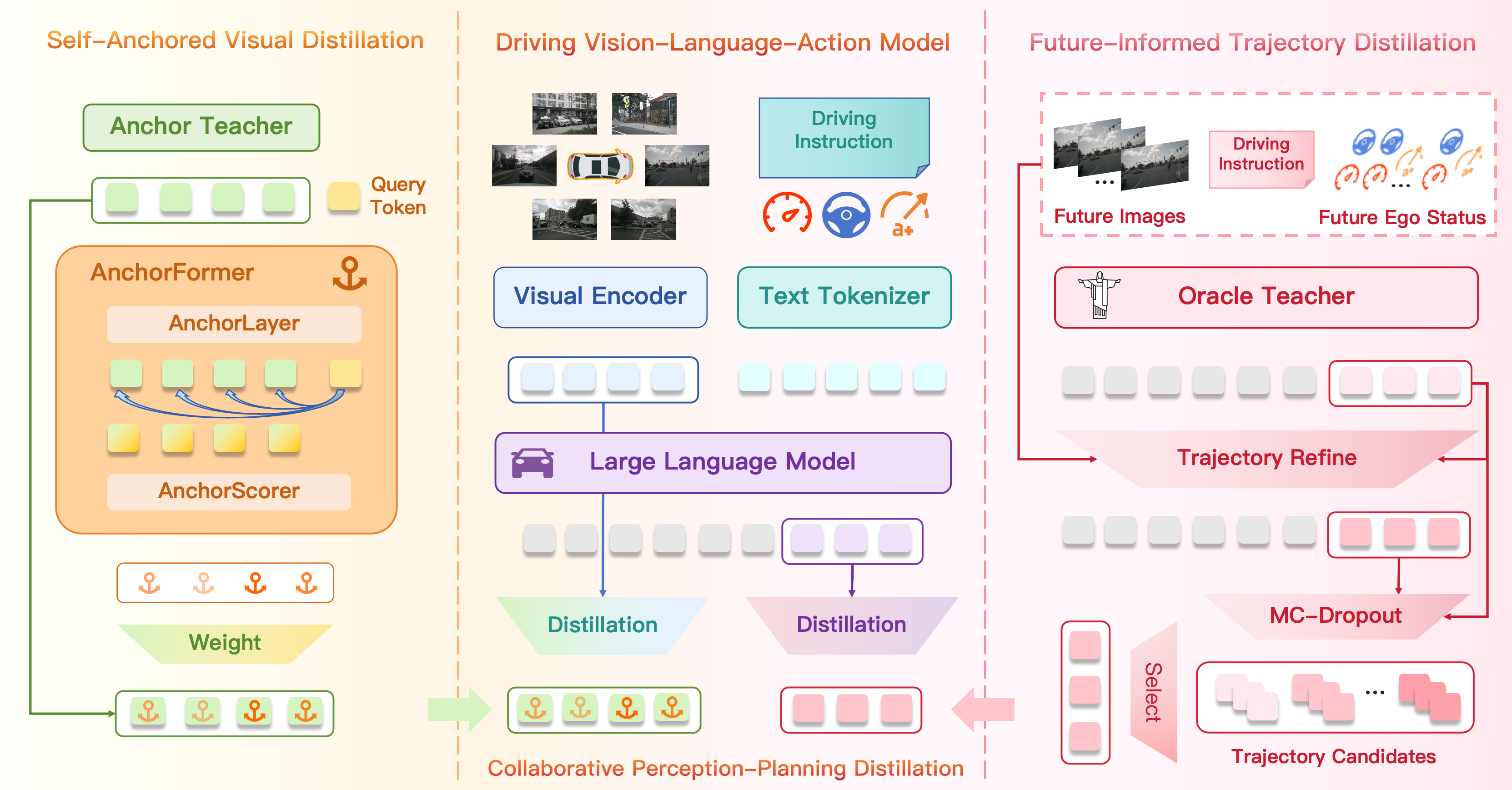}
    \vspace{-0mm}
    \caption{Overview of the EvoDriveVLA framework. \textbf{(Left)} Self-anchored visual distillation  imposes token-leve visual anchoring constraints across the scene; \textbf{(Right)} Future-informed trajectory distillation leverages future ground-truth information for trajectory refinement and diversity sampling; \textbf{(Middle)} Collaborative perception-planning distillation enhances autonomous driving VLA model capabilities in both perception and planning to achieve superior driving performance.}
    \label{fig:main}
    \vspace{-0mm}
\end{figure*}

To address the limitations of existing knowledge distillation methods for VLA-based autonomous driving, we propose EvoDriveVLA, a novel collaborative perception-planning distillation framework incorporating self-anchored and future-informed distillation for autonomous driving. As illustrated in Fig.~\ref{fig:main}, the proposed framework consists of "self-anchored visual distillation" and "future-informed trajectory distillation" to synergistically enhance visual representation and trajectory prediction. Specifically, at the perceptual distillation level, we introduce a self-anchor teacher to provide visual anchoring constraints, preventing the visual encoder from losing its pre-trained representation capabilities after being unfrozen. Simultaneously, trajectory-guided attention is integrated to impose stronger anchoring constraints specifically on critical perceptual regions. At the planning distillation level, we construct a future-aware oracle teacher by incorporating privileged information, including future scene images and ego status, thereby endowing the teacher with superior trajectory prediction accuracy. We further employ a coarse-to-fine trajectory refinement strategy combined with Monte Carlo dropout (MC-Dropout) sampling to synthesize reasoning trajectories that model future evolutions. Subsequently, the optimal trajectory is selected as a soft target for distillation. This enables the student to internalize the oracle's future-predictive capabilities for more robust spatial reasoning and motion prediction. Experimental results demonstrate that EvoDriveVLA achieves leading performance in both open-loop nuScenes~\cite{caesar2020nuscenes} and closed-loop NAVSIM~\cite{dauner2024navsim} evaluations. 

Our contributions are summarized as follows:

\begin{itemize}
\item We propose EvoDriveVLA, a novel collaborative perception-planning distillation framework with self-anchored and future-informed distillation for driving.

\item We introduce self-anchored visual distillation, imposing visual anchoring constraints on trajectory-guided key regions to enhance perceptual capabilities.

\item We propose future-informed trajectory distillation, leveraging an oracle teacher to generate high-quality candidates via trajectory refinement and MC-Dropout, enabling the student to internalize the teacher’s future-aware insights.

\item Our proposed method achieves SOTA performance in open-loop evaluation and significantly enhances performance in closed-loop evaluation.
\end{itemize}

\section{Related Work}
\subsection{End-to-End Autonomous Driving} 
Represented by works such as UniAD~\cite{hu2023planning}, end-to-end methods establish a unified mapping framework from perception to planning, significantly improving the overall adaptability and performance of the system. Furthermore, VAD~\cite{jiang2023vad} and VADv2~\cite{chen2024vadv2} enhance the generalization capability of perception and planning through large-scale visual pre-training, while RAD~\cite{gao2025rad} builds high-fidelity simulation environments based on 3D Gaussian Splatting~\cite{kerbl20233d, wei2025emd, huang2024S3Gaussian, wei2026parkgaussian} for closed-loop optimization. With the rise of generative models, approaches such as DiffusionDrive~\cite{liao2025diffusiondrive} leverage diffusion models to generate high-quality multimodal trajectories, and DiffusionDrivev2~\cite{zou2025diffusiondrivev2} further integrates reinforcement learning to improve prediction diversity and safety in complex scenarios.

\subsection{Vision-Language-Action Models in Driving} Benefiting from the formidable and emergent capabilities of large visual language models (VLMs), Vision–Language–Action (VLA) models~\cite{chi2025impromptu, cao2025fastdrivevla} have increasingly emerged as a promising paradigm in end-to-end autonomous driving. Early works~\cite{sima2024drivelm, tian2024drivevlm} pioneered the use of VLMs for scene understanding via question answering and trajectory planning. Subsequently, OmniDrive~\cite{wang2025omnidrive} and OpenDriveVLA~\cite{zhou2025opendrivevla} further incorporated reasoning tasks and 3D perception modules to improve the accuracy of trajectory prediction and the completeness of environmental modeling. Meanwhile, inspired by the success of reinforcement learning in large language models (LLMs), an increasing number of studies~\cite{li2025recogdrive, guo2025vdrive} are integrating reinforcement learning frameworks into driving VLA models to optimize decision-making.

\subsection{Distilling Knowledge for Autonomous Driving} The success of knowledge distillation in VLMs~\cite{cai2025llava, cao2025move} is gradually being extended to autonomous driving. Early works~\cite{li2024hydra, li2025hydra} focused on traditional end-to-end models, distilling prior knowledge including traffic rules and safety constraints to improve the accuracy and safety of trajectory prediction. DSDrive~\cite{liu2025dsdrive} distills the trajectory outputs of VLMs teacher as soft targets to improve prediction quality, while DistillDrive~\cite{yu2025distilldrive} further introduces a planning vocabulary to diversify teacher-generated trajectories. Furthermore, some studies~\cite{khanzada2025driving} distill dense-reward dynamics from teacher world models into sparse-reward policies for efficient reinforcement learning, while others enhance robustness via cross-modal probabilistic distillation~\cite{liao2025robodrivevlm} or bridge semantic planning and scene understanding through fine-grained feature distillation in diffusion-based planners~\cite{zhang2025lap}.

\section{Methodology}

\subsection{Preliminary}
In autonomous driving, Vision-Language-Action (VLA) models formulate trajectory planning as the prediction of future waypoint sequences conditioned on multi-modal observations. At each time step $t$, the model receives a set of multi-view camera images
$\mathcal{I}_t = \{ I_t^{(v)} \}_{v=1}^{V}$,
a textual instruction prompt $P_t$, and the ego-vehicle state
$S_t = (x_t, y_t, v_t, a_t, \delta_t)$,
which includes vehicle position, velocity, acceleration, and steering angle. The model outputs a sequence of future waypoints
$W_t = \{ w_{t+\tau} \}_{\tau=1}^{T}$,
where each
$w_{t+\tau} = (x_{t+\tau}, y_{t+\tau})$
represents the position at future step $t+\tau$.

% From a modeling perspective, we treat the waypoint sequence $W_t$ as actions and regard multi-view images, instruction prompts, and ego-vehicle states as multi-modal observations
% $\mathcal{O}_t = (\mathcal{I}_t, P_t, S_t)$.

We treat waypoints $W_t$ as actions and observations as $\mathcal{O}_t = (\mathcal{I}_t, P_t, S_t)$. Our goal is to model the conditional distribution of future actions given these observations. Specifically, we aim to learn a policy $p_\theta$ that captures the joint dependencies among the vision, language, and action modalities. During training, we optimize the negative log-likelihood of the predicted distribution with respect to the ground-truth waypoint sequence $W_t^\ast = \{ w^\ast_{t+\tau} \}_{\tau=1}^{T}$, which is defined as the training loss:
\begin{equation}
\mathcal{L}
=
- \sum_{\tau=1}^{T}
\log
p_\theta\!\left(w_{t+\tau} = 
w_{t+\tau}^\ast
\mid
\mathcal{O}_t, w_{<t+\tau-1}^\ast
\right),
\end{equation}
where $\theta$ denotes the learnable model parameters, and $w^\ast_{<t+\tau} = \{w^\ast_{t+1}, \dots, w^\ast_{t+\tau-1}\}$. 
% The training objective is to minimize the loss $\mathcal{L}$.

\subsection{Self-Anchored Visual Distillation}

A long-standing question in VLM research concerns whether the visual encoder should be fully fine-tuned during the SFT. Some studies~\cite{tong2024cambrian, shi2024eagle} argue that unfreezing the visual encoder facilitates cross-domain adaptation and improves visual perception in new domains or downstream tasks. Conversely, others~\cite{karamcheti2024prismatic, kachaev2025don} suggest that fine-tuning may degrade the pre-trained representations, leading to reduced perceptual robustness and overfitting to the training dataset, thereby harming the model’s generalization ability. Therefore, we ask how to enhance task-relevant visual perception for autonomous driving scenarios while preserving the encoder's original capabilities.

\paragraph{Tajectory-Guided Anchoring Constraints.} To address the degradation-adaptation dilemma of visual encoders during supervised fine-tuning, we propose a self-anchored visual distillation. Specifically, we create a self-anchor teacher by copying the student visual encoder before fine-tuning. During training, the stable visual representations produced by this self-anchor teacher are used as distillation constraints, ensuring that the student visual encoder enhances its perception capability for autonomous driving scenarios while preserving its original visual representation power. We further improve granularity by introducing trajectory-guided token-level anchored distillation. To this end, we design AnchorFormer, which assigns adaptive anchor weights to different spatial regions in the scene, where higher weights correlate with intensified anchoring constraints for those regions.

\paragraph{AnchorFormer Architecture.}
AnchorFormer consists of an AnchorLayer and an AnchorScorer. The AnchorLayer shares the same architecture as a single LLM decoder layer, while the AnchorScorer is implemented as a single linear layer. Given multi-view images $I_t$, the self-anchor teacher and student visual encoders produce visual tokens
$\mathbf{z}_v^{tea}$ and $\mathbf{z}_v^{stu}$, respectively. The textual instruction prompt $P_t$, ego-vehicle state $S_t$, and ground-truth future waypoints $W_t^{\ast}$ are encoded into token representations
$\mathbf{z}_{p}$, $\mathbf{z}_{s}$, and $\mathbf{z}_{w^\ast}$. To enable the self-anchor teacher to assign adaptive anchoring weights to visual tokens conditioned on the instruction, ego state, and future trajectory, we introduce a set of learnable query tokens $\mathbf{q}$. These tokens are concatenated with the observation tokens $
\mathbf{z}_{o} = [\mathbf{z}_{v}^{tea}, \mathbf{z}_{p}, \mathbf{z}_{s}]$
and the trajectory tokens $\mathbf{z}_{w^\ast}$, and then fed into the AnchorLayer:
\begin{equation}
\tilde{\mathbf{z}}_{o}, \;
\tilde{\mathbf{z}}_{w^\ast}, \;
\tilde{\mathbf{q}}
=
\mathrm{AnchorLayer}
\left(
\mathbf{z}_{o}, \mathbf{z}_{w^\ast}, \mathbf{q}
\right),
\end{equation}
where
$\tilde{\mathbf{z}}_{o} = (\tilde{\mathbf{z}}_{v}^t, \tilde{\mathbf{z}}_{p}, \tilde{\mathbf{z}}_{s})$. We compute token-level anchor scores by applying the AnchorScorer to the Hadamard product between the updated visual tokens $\tilde{\mathbf{z}}_{v}^t$ and query tokens $\tilde{\mathbf{q}}$:
\begin{equation}
\mathbf{S}_a
=
\mathrm{AnchorScorer}
\left(
\tilde{\mathbf{z}}_{v}^t \odot \tilde{\mathbf{q}}
\right).
\end{equation}
The anchor weights are subsequently obtained via a temperature-scaled sigmoid normalization:
\begin{equation}
\mathbf{W}_a
=
\frac{1}{1 + \exp\!\left(-\mathbf{S}_a / \tau_v \right)},
\end{equation}
where the temperature is set to $\tau_v = 2.0$.

\paragraph{Visual Distillation Loss.}
We adopt a mean squared error (MSE) loss to constrain the student's visual tokens $\mathbf{z}_v^{stu}$ with the self-anchor teacher's visual tokens $\mathbf{z}_v^{tea}$, weighted by the token-level anchor weights $\mathbf{W}_a$. The resulting self-anchored distillation loss $\mathcal{L}_a$ is defined as:
\begin{equation}
\mathcal{L}_a
=
\frac{1}{N_v}
\sum_{i=1}^{N_v}
\mathbf{W}_a^{(i)}
\,
\left\|
\mathbf{z}_{v}^{tea(i)} - \mathbf{z}_{v}^{stu(i)}
\right\|_2^2,
\end{equation}

where $N_v$ denotes the number of visual tokens.

\subsection{Future-Informed Trajectory Distillation}

\paragraph{The Future-Aware Oracle Teacher.}

We contend that the bottleneck of trajectory prediction in autonomous driving lies in the model's capacity to anticipate complex scene evolutions and potential agent interactions. Existing distillation paradigms~\cite{liu2025dsdrive, hegde2025distilling} typically employ teachers that share the same observation space as the student. Consequently, these teachers fail to provide insights beyond the student's own perceptual limits, offering little more than saturated guidance. To transcend this limitation, we introduce an oracle teacher with access to privileged future information. 

By conditioning on observations from the subsequent $T$ seconds ($\mathcal{O}_{t+1}, \dots, \mathcal{O}_{t+T-1}$), the teacher can capture the actual spatio-temporal evolution of the scene, establishing a stronger temporal upper bound that surpasses the student’s perceptual horizon. The essence of our future-informed trajectory distillation lies in enabling the student model to internalize the future-aware insights encapsulated within the oracle teacher. Crucially, this mechanism preserves a fair evaluation setting, as the student remains strictly confined to current observations during both training and inference. To further bolster the teacher's predictive quality, we implement a progressive coarse-to-fine refinement strategy:

\vspace{-0mm}

\begin{equation}
\begin{aligned}
p_\theta(W_t^{c}\mid\cdot)
&=\prod_{\tau=1}^{T}
p_\theta\!\left(w_{t+\tau}\mid\mathcal{O}_{<t+T},\,w_{<t+\tau}^\ast\right),\\
p_\theta(W_t^{f}\mid\cdot)
&=\prod_{\tau=1}^{T}
p_\theta\!\left(w_{t+\tau}\mid\mathcal{O}_{<t+T},\,W_t^{c},\,w_{<t+\tau}^\ast\right),
\end{aligned}
\end{equation}

where $W_t^{c}$ and $W_t^{f}$ denote coarse and refined trajectory sequences, respectively. During training, a joint sampling-based strategy is employed to optimize both schemes, empowering the teacher model to simultaneously capture both coarse and fine-grained trajectory representations.

\paragraph{Coarse-to-Fine Trajectory Refinement.} 
We feed the oracle teacher's coarse trajectories back into the model for iterative refinement. Leveraging future-aware perception, the oracle teacher rectifies candidate trajectories for spatio-temporal consistency, yielding smoother and physically plausible optimized paths. This recursive generate-refine process effectively simulates the progressive trajectory evolution under oracle guidance. To help the student inherit these future-informed priors, we include hidden states and logits from both coarse and fine-grained trajectories in the candidate sets $\mathcal{S}_h = \left\{ \mathbf{h}_{\text{c}}, \mathbf{h}_{\text{f}} \right\}$ and $\mathcal{S}_l = \left\{ \mathbf{l}_{\text{c}}, \mathbf{l}_{\text{f}} \right\}$, respectively. This strategy allows the student to effectively internalize the nuanced corrective capabilities and future-oriented insights of the oracle teacher.

\begin{table*}[t]
  \centering
  \caption{Open-loop evaluation on nuScenes. We conduct open-loop evaluation on the nuScenes, comparing our method against traditional, LLM-based, and distillation-based baselines.}
  \label{tab:main}
  \renewcommand{\arraystretch}{1.1} 
  \resizebox{\linewidth}{!}{
    \begin{tabular}{l cccc cccc | cccc cccc}
    \toprule
    \multirow{3}{*}{Method} & \multicolumn{8}{c}{ST-P3} & \multicolumn{8}{c}{UniAD} \\ 
    \cmidrule(lr){2-9} \cmidrule(lr){10-17}
    & \multicolumn{4}{c}{L2 (m) $\downarrow$} & \multicolumn{4}{c}{Collision (\%) $\downarrow$} & \multicolumn{4}{c}{L2 (m) $\downarrow$} & \multicolumn{4}{c}{Collision (\%) $\downarrow$} \\
    \cmidrule(lr){2-5} \cmidrule(lr){6-9} \cmidrule(lr){10-13} \cmidrule(lr){14-17}
    & 1s & 2s & 3s & Avg. & 1s & 2s & 3s & Avg. & 1s & 2s & 3s & Avg. & 1s & 2s & 3s & Avg.\\
    
    \rowcolor{lightgray!50}\multicolumn{17}{c}{\textit{Traditional}} \\

    \rowcolor{red!10} ST-P3 \footnotesize{[ECCV 22]} & 1.44 & 2.11 & 2.90 & 2.11 & 0.23 & 0.62 & 1.27 & 0.71 & - & - & - & - & - & - & - & - \\

    \rowcolor{red!10} VAD \footnotesize{[ICCV 23]} & 0.17 & 0.34 & 0.60 & 0.37 & 0.04 & 0.27 & 0.67 & 0.33 & - & - & - & - & - & - & - & - \\

    \rowcolor{red!10} BEV-Planner \footnotesize{[CVPR 24]} & 0.16 & 0.32 & 0.57 & 0.35 & 0.00 & 0.29 & 0.73 & 0.34 & - & - & - & - & - & - & - & - \\

    \rowcolor{red!10} DiffusionDrive \footnotesize{[CVPR 25]} & 0.27 & 0.54 & 0.90 & 0.57 & 0.03 & 0.05 & 0.16 & 0.08 & - & - & - & - & - & - & - & - \\
    
    \rowcolor{red!10} FF \footnotesize{[CVPR 21]} & - & - & - & - & - & - & - & - & 0.55 & 1.20 & 2.54 & 1.43 & 0.06 & 0.17 & 1.07 & 0.43 \\
    
    \rowcolor{red!10} EO \footnotesize{[ECCV 22]} & - & - & - & - & - & - & - & - & 0.67 & 1.36 & 2.78 & 1.60 & 0.04 & 0.09 & 0.88 & 0.33 \\

    \rowcolor{red!10} PreWorld \footnotesize{[ICLR 25]} & - & - & - & - & - & - & - & - & 0.49 & 1.22 & 2.32 & 1.34 & 0.19 & 0.57 & 2.65 & 1.14 \\

    \rowcolor{red!10} UniAD \footnotesize{[CVPR 23]} & 0.44 & 0.67 & 0.96 & 0.69 & 0.04 & 0.08 & 0.23 & 0.12 & 0.48 & 0.96 & 1.65 & 1.03 & 0.05 & 0.17 & 0.71 & 0.31 \\

    \rowcolor{lightgray!50}\multicolumn{17}{c}{\textit{LLM-Based}} \\

    \rowcolor{green!10} DriveVLM \footnotesize{[CoRL 24]} & 0.18 & 0.34 & 0.68 & 0.40 & 0.10 & 0.22 & 0.45 & 0.27 & - & - & - & - & - & - & - & - \\
    
    \rowcolor{green!10} OmniDrive \footnotesize{[CVPR 25]} & 0.14 & 0.29 & 0.55 & 0.33 & 0.00 & 0.13 & 0.78 & 0.30 & - & - & - & - & - & - & - & - \\
    
    \rowcolor{green!10} ORION \footnotesize{[ICCV 25]} & 0.17 & 0.31 & 0.55 & 0.34 & 0.05 & 0.25 & 0.80 & 0.37 & - & - & - & - & - & - & - & - \\

    \rowcolor{green!10} ELM \footnotesize{[ECCV 24]} & - & - & - & - & - & - & - & - & 0.34 & 1.23 & 2.57 & 1.38 & 0.12 & 0.50 & 2.36 & 0.99 \\
    
    % \rowcolor{green!10} FeD \footnotesize{[CVPR 24]} & - & - & - & - & - & - & - & - & 0.27 & 0.53 & 0.94 & 0.58 & 0.00 & 0.04 & 0.52 & 0.19 \\
        
    \rowcolor{green!10} DME-Driver \footnotesize{[AAAI 25]} & - & - & - & - & - & - & - & - & 0.45 & 0.91 & 1.58 & 0.98 & 0.05 & 0.28 & 0.55 & 0.29 \\

    \rowcolor{green!10} GPT-Driver \footnotesize{[NeurIPS 23]} & 0.20 & 0.40 & 0.70 & 0.44 & 0.04 & 0.12 & 0.36 & 0.17 & 0.27 & 0.74 & 1.52 & 0.84 & 0.07 & 0.15 & 1.10 & 0.44 \\

    \rowcolor{green!10} OccWorld \footnotesize{[ECCV 24]} & 0.39 & 0.73 & 1.18 & 0.77 & 0.11 & 0.19 & 0.67 & 0.32 & 0.52 & 1.27 & 2.41 & 1.40 & 0.12 & 0.40 & 2.08 & 0.87 \\

    \rowcolor{green!10} OpenDriveVLA \footnotesize{[AAAI 26]} & 0.14 & 0.30 & 0.55 & 0.33 & 0.02 & 0.07 & 0.22 & 0.10 & 0.19 & 0.58 & 1.24 & 0.67 & 0.02 & 0.18 & 0.70 & 0.30 \\

    \rowcolor{lightgray!50}\multicolumn{17}{c}{\textit{Distillation-Based}} \\

    \rowcolor{cyan!10} DistillDrive \footnotesize{[ICCV 25]} & 0.28 & 0.54 & 0.83 & 0.57 & 0.00 & 0.03 & 0.17 & 0.06 & - & - & - & - & - & - & - & - \\

    \rowcolor{cyan!10} DiMA \footnotesize{[CVPR 25]} & 0.12 & 0.25 & 0.44 & 0.27 & 0.04 & 0.06 & 0.15 & 0.08 & 0.18 & 0.48 & 1.01 & 0.57 & 0.00 & 0.05 & 0.16 & 0.07 \\

    \rowcolor{blue!10} \textbf{EvoDriveVLA(Ours)} & \textbf{0.12} & \textbf{0.24} & \textbf{0.43} & \textbf{0.26} & \underline{0.02} & \underline{0.05} & \textbf{0.12} & \textbf{0.06} & \textbf{0.16} & \textbf{0.44} & \textbf{0.96} & \textbf{0.52} & \underline{0.02} & \textbf{0.02} & \underline{0.33} & \underline{0.12} \\
    
    \bottomrule
    \end{tabular}
  }
  \vspace{0mm}
\end{table*}

\paragraph{MC-Dropout Trajectory Sampling.} 
Although the coarse-to-fine strategy enables the oracle teacher to rectify its predictions, we further introduce a Monte Carlo Dropout (MC-Dropout) trajectory sampling strategy to enhance both the precision and robustness of the supervision signal. Instead of relying on a single deterministic forward pass, we apply $N$ stochastic dropout perturbations to the oracle teacher's output hidden state $\mathbf{h} \in \mathcal{S}_h$ while keeping parameters fixed, resulting in a set of diversified hidden state samples:
\begin{equation}
\mathbf{h}^{(n)} = \mathrm{Dropout}\!\left( \mathbf{h}; \, p \right),
\quad n = 1, \dots, N,
\end{equation}
where $p=0.1$ denotes the dropout rate and $N=10$. The sampled hidden states are then fed into the model's $\mathbf{lm\_head}$ to obtain the corresponding logits. Finally, all sampled hidden states and their associated logits are incorporated into the candidate sets $\mathcal{S}_h$ and $\mathcal{S}_l$, respectively:

\begin{equation}
\mathcal{S}_h \leftarrow \mathcal{S}_h \cup \left\{ \mathbf{h}^{(n)} \right\}_{n=1}^{N}, \quad
\mathcal{S}_l \leftarrow \mathcal{S}_l \cup \left\{ \mathbf{l}^{(n)} = \operatorname{lm\_head}\left( \mathbf{h}^{(n)} \right) \right\}_{n=1}^{N}.
\end{equation}

Since MC-Dropout is applied only to hidden states and logits are computed through the lightweight $\mathbf{lm\_head}$, this strategy incurs minimal overhead while enabling the oracle teacher to identify and select the most reliable trajectory paths from an expanded search space.

\paragraph{Trajectory Distillation Loss.} 
We compute the cross-entropy loss between predicted logits in $\mathcal{S}_l$ and the ground-truth, and select the optimal trajectory with the minimum loss:

\begin{equation}
\hat{k} = 
\arg\min_{ \mathbf{l}_k \in \mathcal{S}_l }
\;\mathcal{L}_{\mathrm{CE}}\!\left(
\mathbf{l}_k,\,
\mathbf{W}^\ast
\right).
\end{equation}

We then distill the student model using the hidden states and logits associated with this optimal future-informed trajectory as soft targets. This encourages the student to align with the oracle teacher's privileged knowledge in both the latent representation space and the predictive distribution. The formulation is given as follows:

\begin{equation}
\mathcal{L}_{\mathrm{h}} = \frac{1}{N_t} \sum_{i=1}^{N_t} \left\| \mathbf{h}_{stu}^{(i)} - \mathbf{h}_{\hat{k}}^{(i)} \right\|_2^2, 
\quad 
\mathcal{L}_{\mathrm{l}} = \mathrm{KL}\left( \mathrm{softmax}(\mathbf{l}_{\hat{k}}/\tau_t) \;\|\; \mathrm{softmax}(\mathbf{l}_{stu}/\tau_t) \right)
\end{equation}

where $\tau_t=5$, $N_t$ denotes the number of trajectory tokens, $\mathbf{h}_{stu}$ and $\mathbf{l}_{stu}$ are student's hidden states and logist, respectively. Essentially, this dual-level alignment enables the student to internalize the reasoning required for trajectory prediction.

\subsection{Overall Training Loss}
The overall training loss $\mathcal{L}_{all}$ of the student model is a weighted combination composed of the trajectory prediction loss $\mathcal{L}$, the self-anchored visual distillation loss $\mathcal{L}_a$, and the future-informed trajectory distillation loss terms $\mathcal{L}_{\mathrm{h}}$ and $\mathcal{L}_{\mathrm{l}}$, which can be formulated as follows:
\vspace{-1mm}
\begin{equation}
\mathcal{L}_{all} = \mathcal{L} + \lambda_a * \mathcal{L}_a + \lambda_h * \mathcal{L}_h + \lambda_l * \mathcal{L}_l ,
\end{equation}
where we set $\lambda_a = 0.05$, $\lambda_h = 0.1$, and $\lambda_l = 0.2$.

\begin{table*}[t!]
    \centering
    \small
    \caption{Performance comparison on NAVSIM \textit{navtest} using closed-loop metrics.}
    
    \setlength{\tabcolsep}{8pt}
    \begin{tabular}{l|cc|ccc|cc}
        \toprule
        Method & NC$\uparrow$ & DAC$\uparrow$ & TTC$\uparrow$ & Comf. $\uparrow$ & EP$\uparrow$ & PDMS$\uparrow$ \\
        \midrule
        
        Constant Velocity & 68.0 & 57.8 & 50.0 & 100 & 19.4 & \cellcolor{gray!30} 20.6 \\
        
        Ego Status MLP & 93.0 & 77.3 & 83.6 & 100 & 62.8 & \cellcolor{gray!30} 65.6 \\        
        
        \midrule
        
        VADv2-$\mathcal{V}_{\text{8192}}$ \footnotesize{[ICCV 23]} & 97.2 & 89.1 & 91.6 & 100 & 76.0 &\cellcolor{gray!30}  80.9 \\
        
        % DrivingGPT & \checkmark &  &  \textbf{98.9} & 90.7 & 94.9 & 95.6 & 79.7 & \cellcolor{gray!30} 82.4 \\ 
        
        UniAD \footnotesize{[CVPR 23]} & 97.8 & 91.9 & 92.9 & 100 & 78.8 & \cellcolor{gray!30} 83.4 \\
        
        TransFuser \footnotesize{[TPAMI 23]} & 97.7 & 92.8 & 92.8 & 100 & 79.2 & \cellcolor{gray!30} 84.0 \\
        
        PARA-Drive \footnotesize{[CVPR 24]} & 97.9 & 92.4 & 93.0 & 99.8 & 79.3 & \cellcolor{gray!30} 84.0 \\
        
        % DRAMA & \checkmark & \checkmark   & 98.0 & 93.1 & 94.8 & 100 & 80.1 & \cellcolor{gray!30} 85.5 \\
        
        % Hydra-MDP-$\mathcal{V}_{\text{8192}}$-W-EP & \checkmark & \checkmark  & 98.3 & 96.0 & 94.6 & \textbf{100} & 78.7 & \cellcolor{gray!30} 86.5 \\
        
        % DiffusionDrive & \checkmark & \checkmark & 98.2 & 96.2 & 94.7 & \textbf{100} & 82.2 & \cellcolor{gray!30} 88.1 \\
        
        % WoTE & \checkmark & \checkmark & 98.5 & 96.8 & \textbf{94.9} & 99.9 & 81.9 & \cellcolor{gray!30} 88.3 \\
        
        \midrule
        QwenVL2.5-3B \footnotesize{[arXiv 25]} & 97.3 & 90.4 & 92.9 & 99.6 & 77.6 & \cellcolor{gray!30} 81.9 \\
        
        QwenVL2.5-8B \footnotesize{[arXiv 25]} & 97.8 & 92.1 & 92.8 & 100 & 78.3 & \cellcolor{gray!30} 83.3 \\
        
        InternVL3-8B \footnotesize{[arXiv 25]} & 97.0 & 92.4 & 91.8 & 100 & 78.9 & \cellcolor{gray!30} 83.3 \\

        \textbf{EvoDriveVLA(Ours)} & \textbf{98.0} & \textbf{93.3} & \textbf{93.1} & \textbf{100} & \textbf{81.1} & \cellcolor{gray!30} \textbf{85.3} \\
        \bottomrule
    \end{tabular}
    \label{tab:navsim}
\end{table*}

% \usepackage{amssymb} % 提供 \checkmark
% \usepackage{xcolor}

% 定义稳健的对号和错号
% \checkmark 是原生的对号
% \times 是数学符号中的叉号，通过 \mathbin 调整间距
\newcommand{\cmark}{\textcolor{green!80!black}{\checkmark}} 
\newcommand{\xmark}{\textcolor{red!80!black}{${\times}$}}

\begin{table}[t]
  \centering
  \caption{Ablation study on algorithmic components.}
  \label{tab:ablation}
  % 1. 增加行高
  \renewcommand{\arraystretch}{1.1} 
  % 2. 减小列间距，防止表格溢出
  \setlength{\tabcolsep}{4pt} 
  % 3. 使用标准字号命令，而不是强制缩放
  \small 
  
  \begin{tabular}{ccccc|cccc|cccc}
    \toprule
    \multicolumn{5}{c|}{Ablation} & \multicolumn{4}{c}{L2 (m) $\downarrow$} & \multicolumn{4}{c}{Collision (\%) $\downarrow$} \\  

    \cmidrule(lr){1-5} \cmidrule(lr){6-9} \cmidrule(lr){10-13}

    Traj KD & Refine & MC-Drop & Frozen Enc & Visual KD & 1s & 2s & 3s & Avg. & 1s & 2s & 3s & Avg. \\

    \cmidrule(lr){1-5} \cmidrule(lr){6-9} \cmidrule(lr){10-13}
    
    \xmark & \xmark & \xmark & \xmark & \xmark & 0.17 & 0.47 & 1.02 & 0.55 & 0.03 & 0.08 & 0.45 & 0.19 \\
    \cmark & \xmark & \xmark & \xmark & \xmark & 0.17 & 0.46 & 1.00 & 0.54 & 0.05 & 0.07 & 0.38 & 0.17\\
    \cmark & \cmark & \xmark & \xmark & \xmark & 0.16 & 0.46 & 0.99 & 0.53 & 0.05 & 0.03 & 0.40 & 0.16\\
    \cmark & \cmark & \cmark & \xmark & \xmark & 0.16 & 0.45 & 0.98 & 0.53 & 0.03 & 0.05 & 0.35 & 0.14\\
    \cmark & \cmark & \cmark & \cmark & \xmark & 0.17 & 0.47 & 1.04 & 0.56 & 0.03 & 0.10 & 0.42 & 0.18 \\
    \cmark & \cmark & \cmark & \xmark & \cmark & \textbf{0.16} & \textbf{0.44} & \textbf{0.96} & \textbf{0.52} & \textbf{0.02} & \textbf{0.02} & \textbf{0.33} & \textbf{0.12}\\

    \bottomrule
  \end{tabular}
  \vspace{-2mm}
\end{table}

\section{Experiments}
\subsection{Experimental Settings} 
\paragraph{Implementation Details.} Both the student model and the oracle teacher share the same Qwen2.5-VL 3B~\cite{bai2025qwen2} architecture, while the corresponding visual encoder serves as the self-anchor teacher. Furthermore, the AnchorLayer is initialized with the weights from its final LLM layer. During the distillation training process
, the weights of both the oracle teacher and the self-anchor teacher remain frozen, while only the parameters of the student model and the AnchorFormer are actively optimized.

\paragraph{Datasets and Evaluations.} We evaluate open-loop performance on nuScenes~\cite{caesar2020nuscenes} following ST-P3~\cite{hu2022st} and UniAD~\cite{hu2023planning} protocols, measuring L2 errors and collision rates. For closed-loop assessment, we use the NAVSIM~\cite{dauner2024navsim} benchmark. We report the PDM-Score (PDMS), a composite metric including No Collision (NC), Drivable Area Compliance (DAC), Time to Collision (TTC), Comfort (Comf.), and Ego Progress (EP), alongside planning performance over a 4-second horizon.
% For open-loop evaluation, we utilize the nuScenes benchmark~\cite{caesar2020nuscenes}, which comprises 1,000 driving scenes, each lasting approximately 20 seconds. The dataset is partitioned into training and validation sets according to the standard data split. Our evaluation protocol strictly follows the settings established by both ST-P3~\cite{hu2022st} and UniAD~\cite{hu2023planning}. The performance is measured using L2 displacement errors at 1, 2, and 3-second intervals, along with the average collision rate throughout the prediction horizon.

% For closed-loop evaluation, we employ the NAVSIM benchmark~\cite{dauner2024navsim}. The dataset is partitioned into navtrain (1,192 training scenes) and navtest (136 evaluation scenes). We adopt the PDM-Score (PDMS) as the primary evaluation metric, which provides a comprehensive assessment through several sub-metrics: No Collision (NC), Drivable Area Compliance (DAC), Time to Collision (TTC), Comfort (Comf.), and Ego Progress (EP). Furthermore, we evaluate and report the planning performance over a 4-second prediction horizon.

\begin{figure}[t]
    \centering
    \includegraphics[width=0.99\textwidth]{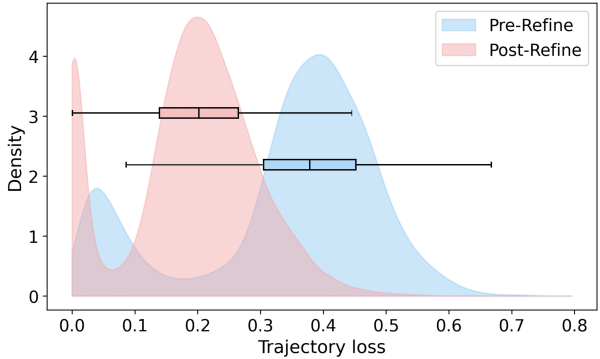}
    \vspace{-0mm}
    \caption{\textbf{(Left)}: Kernel density estimation of trajectory loss distributions for pre-refine and post-refine trajectories. The overlaid boxplots summarize the median, interquartile range, and extreme values. \textbf{(Right)}: Comparison of trajectory loss distributions before and after MC-Dropout sampling.}
    \label{fig:refine_and_mc_dropout}
    \vspace{-0mm}
\end{figure}

% \begin{figure}[t]
%     \centering
%     \includegraphics[width=0.48\textwidth]{figures/traj_mcdrop.png}
%     \vspace{-0mm}
%     \caption{Comparison of trajectory loss distributions before and after MC-Dropout trajectory sampling.}
%     \label{fig:traj_mcdrop}
%     \vspace{-0mm}
% \end{figure}
 
\subsection{Open-loop Evaluation} 
We evaluate the open-loop trajectory planning performance on the nuScenes~\cite{caesar2020nuscenes} benchmark. Specifically, we compare our method against three categories of baselines: traditional~\cite{hu2022st, jiang2023vad, li2024ego, liao2025diffusiondrive, hu2021safe, khurana2022differentiable, li2025semi, hu2023planning}, LLM-based~\cite{tian2024drivevlm, wang2025omnidrive, fu2025orion, zhou2024embodied, han2025dme, mao2023gpt, zheng2024occworld,zhou2025opendrivevla}, and distillation-based~\cite{yu2025distilldrive, hegde2025distilling} approaches. As illustrated in Tab.~\ref{tab:main}, our method achieves the state of the art performance across all three categories, significantly outperforming both traditional and LLM-based baselines by a substantial margin. Compared to OpenDriveVLA, our method achieves significant performance gains: specifically, we improve L2 error and Collision rate by 21\% and 40\% under the ST-P3 setting, and by 22\% and 60\% under the UniAD protocol, respectively. 
% Our approach consistently outperforms distillation-based methods across nearly all metrics.
% Among knowledge distillation-based methods, only DiMA shows a marginal advantage in the collision metric under the UniAD evaluation protocol. Nevertheless, our approach remains significantly superior across all other evaluation dimension, achieving 9\% improvement in L2 error rate over it under the UniAD setting.

% \usepackage{wrapfig}

% \begin{wraptable}{r}{0.48\columnwidth} % r=右侧
%   \centering
%   \caption{Oracle teacher performance on nuScenes.}
%   \label{tab:teacher}
%   \resizebox{\linewidth}{!}{
%     \begin{tabular}{l cccc cccc}
%       \toprule
%       \multirow{2}{*}{Metric} 
%         & \multicolumn{4}{c}{L2 (m) $\downarrow$} 
%         & \multicolumn{4}{c}{Collision (\%) $\downarrow$} \\
%       \cmidrule(lr){2-5} \cmidrule(lr){6-9}
%         & 1s & 2s & 3s & Avg. & 1s & 2s & 3s & Avg. \\
%       \midrule
%       ST-P3 & 0.10 & 0.14 & 0.18 & 0.14 & 0.02 & 0.03 & 0.05 & 0.04 \\
%       UniAD & 0.13 & 0.20 & 0.27 & 0.20 & 0.02 & 0.05 & 0.05 & 0.04 \\
%       \bottomrule
%     \end{tabular}
%   }
% \end{wraptable}

\begin{table}[t]
  \centering
  \caption{Oracle teacher performance on nuScenes.}
  \label{tab:teacher}

  \resizebox{0.7\linewidth}{!}{
    \begin{tabular}{l cccc cccc}
      \toprule
      \multirow{2}{*}{Metric} 
        & \multicolumn{4}{c}{L2 (m) $\downarrow$} 
        & \multicolumn{4}{c}{Collision (\%) $\downarrow$} \\
      \cmidrule(lr){2-5} \cmidrule(lr){6-9}
        & 1s & 2s & 3s & Avg. & 1s & 2s & 3s & Avg. \\
      \midrule
      ST-P3 & 0.10 & 0.14 & 0.18 & 0.14 & 0.02 & 0.03 & 0.05 & 0.04 \\
      UniAD & 0.13 & 0.20 & 0.27 & 0.20 & 0.02 & 0.05 & 0.05 & 0.04 \\
      \bottomrule
    \end{tabular}
  }
    \vspace{0mm}
\end{table}

\begin{figure*}[t]
    \centering
    \includegraphics[width=0.99\textwidth]{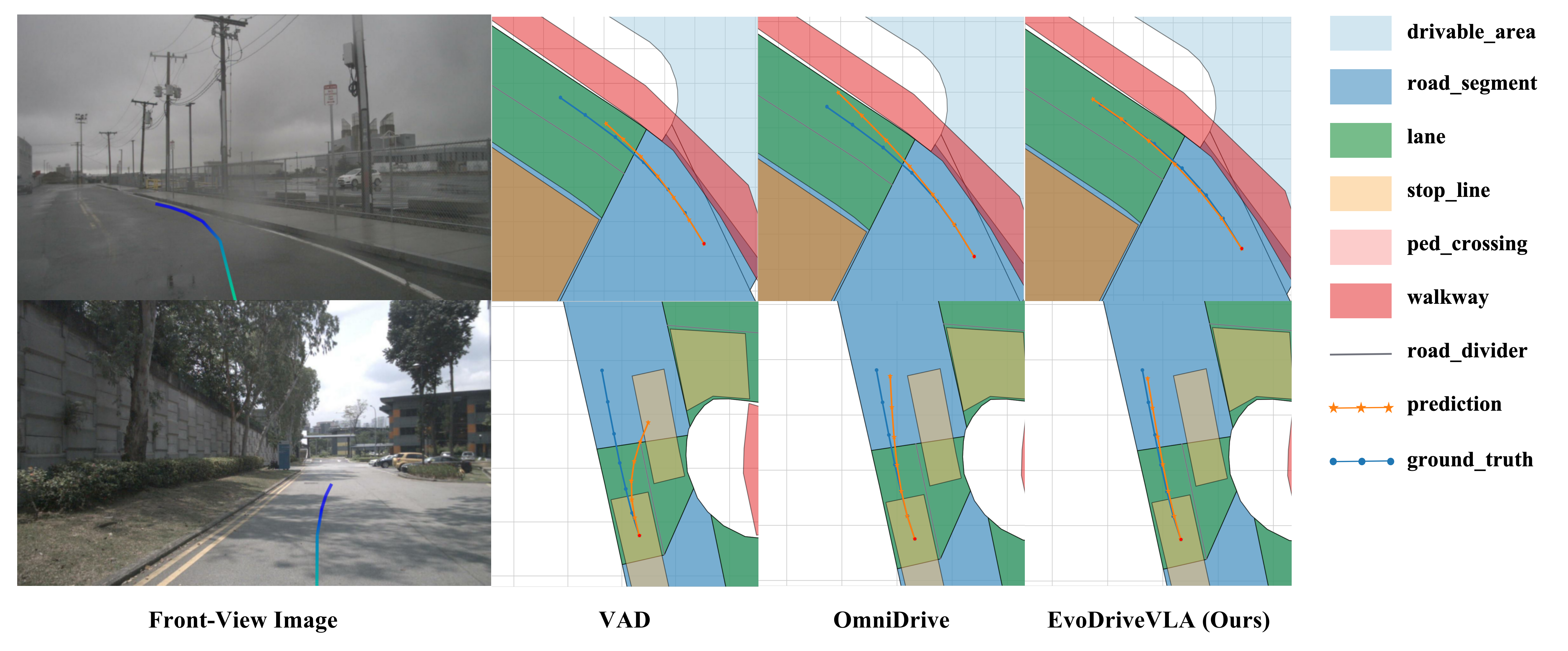}
    \vspace{-0mm}
    \caption{Qualitative comparison on nuScenes.}
    \label{fig:visual_compare}
    \vspace{-2mm}
\end{figure*}

\subsection{Close-loop Evaluation} 
In the closed-loop evaluation, we compared our approach with other camera-only methods~\cite{chen2024vadv2, hu2023planning, chitta2022transfuser, weng2024drive} on the NAVSIM benchmark. As shown in Tab.~\ref{tab:navsim}, our method achieves SOTA performance among these competitors. Additionally, we introduced the 3B and 8B versions of Qwen2.5-VL~\cite{bai2025qwen2}, alongside InternVL3-8B~\cite{zhu2025internvl3} as baselines. Experimental results demonstrate that our distillation algorithm improves the 3B base model's PDMS by 3.4 points (4.2\%). Remarkably, the distilled 3B model outperforms larger models like Qwen2.5-VL 8B and InternVL3-8B by 2.0 points (2.4\%), underscoring its effectiveness in enhancing closed-loop driving performance.

% Experimental results demonstrate that our proposed distillation algorithm improves the PDMS score of the 3B base model by 3.4 points (a 4.2\% increase). Remarkably, the distilled 3B model even outperforms larger-scale models such as Qwen2.5-VL 8B and InternVL3-8B, achieving a 2.0-point lead (a 2.4\% increase) in PDMS. These results underscore the effectiveness of our distillation approach in enhancing the model's closed-loop driving performance.

\begin{table*}[t]
  \centering
  \caption{Ablation on MC-Dropout trajectory sampling.}
  \label{tab:MC-Dropout}
  
  \resizebox{0.8\linewidth}{!}{
    \begin{tabular}{l cccc cccc}
    \toprule
    \multirow{2}{*}{Methods}
    & \multicolumn{4}{c}{L2 (m) $\downarrow$} & \multicolumn{4}{c}{Collision (\%) $\downarrow$} \\
    \cmidrule(lr){2-5} \cmidrule(lr){6-9}
    & 1s & 2s & 3s & Avg. & 1s & 2s & 3s & Avg. \\
    \cmidrule(lr){1-9}
    w/o & 0.16 & 0.46 & 0.99 & 0.53 & 0.05 & 0.03 & 0.40 & 0.16 \\

    Parameter & 0.17 & 0.46 & 1.01 & 0.54 & 0.07 & 0.07 & 0.38 & 0.17 \\

    Hidden State & 0.16 & 0.44 & 0.97 & 0.52 & 0.03 & 0.03 & 0.30 & 0.12 \\
    
    \bottomrule
    \end{tabular}
  }
  \vspace{-4mm}
\end{table*}

\subsection{Ablation Study} 

We conduct ablation studies on the nuScenes benchmark using UniAD metrics, with results summarized in Tab.~\ref{tab:ablation}. The results show that future-informed trajectory distillation significantly enhances prediction accuracy. This improvement is attributed to the superior trajectory prediction capability of the oracle teacher, as well as the coarse-to-fine refinement and MC-Dropout sampling strategies. As illustrated in Tab.~\ref{tab:teacher}, it is evident that the oracle teacher, empowered by future information input, significantly outperforms existing methods in terms of both L2 error and collision rate. Meanwhile, self-anchored visual distillation imposes constraints on the student model’s original perceptual representations. Compared to frozen and unfrozen encoder baselines, our visual distillation approach effectively enhances perception for autonomous driving.

We statistically analyze the loss distribution between teacher-predicted trajectories and ground truth before and after the coarse-to-fine trajectory refinement, visualized via kernel density estimation (KDE) plots. As illustrated in Fig.~\ref{fig:refine_and_mc_dropout} (left), the refinement process causes the trajectory loss distribution to shift significantly toward the lower-value region. Notably, the density near zero markedly increases, while the long-tail distribution of outliers is substantially alleviated. These observations demonstrate the effectiveness of coarse-to-fine refinement in enhancing teacher trajectory prediction.

Furthermore, we analyze teacher trajectory loss variation before and after MC-Dropout sampling. As shown in Fig.~\ref{fig:refine_and_mc_dropout} (right), this strategy further reduces the teacher's prediction error, providing the student with more precise trajectory guidance. Notably, the loss in the near-zero region drops by approximately 50\%, with nearly 30\% of teacher trajectories achieving an L2 loss below 0.1. We also investigate the placement of MC-Dropout, as shown in Tab.~\ref{tab:MC-Dropout}. Applying MC-Dropout to model parameters degrades performance; however, performing it on hidden states not only boosts accuracy but also enables multi-trajectory sampling with negligible latency.

\begin{wrapfigure}{r}{0.50\columnwidth} % r=右侧，l=左侧
    \centering
    \vspace{-5mm} \includegraphics[width=\linewidth]{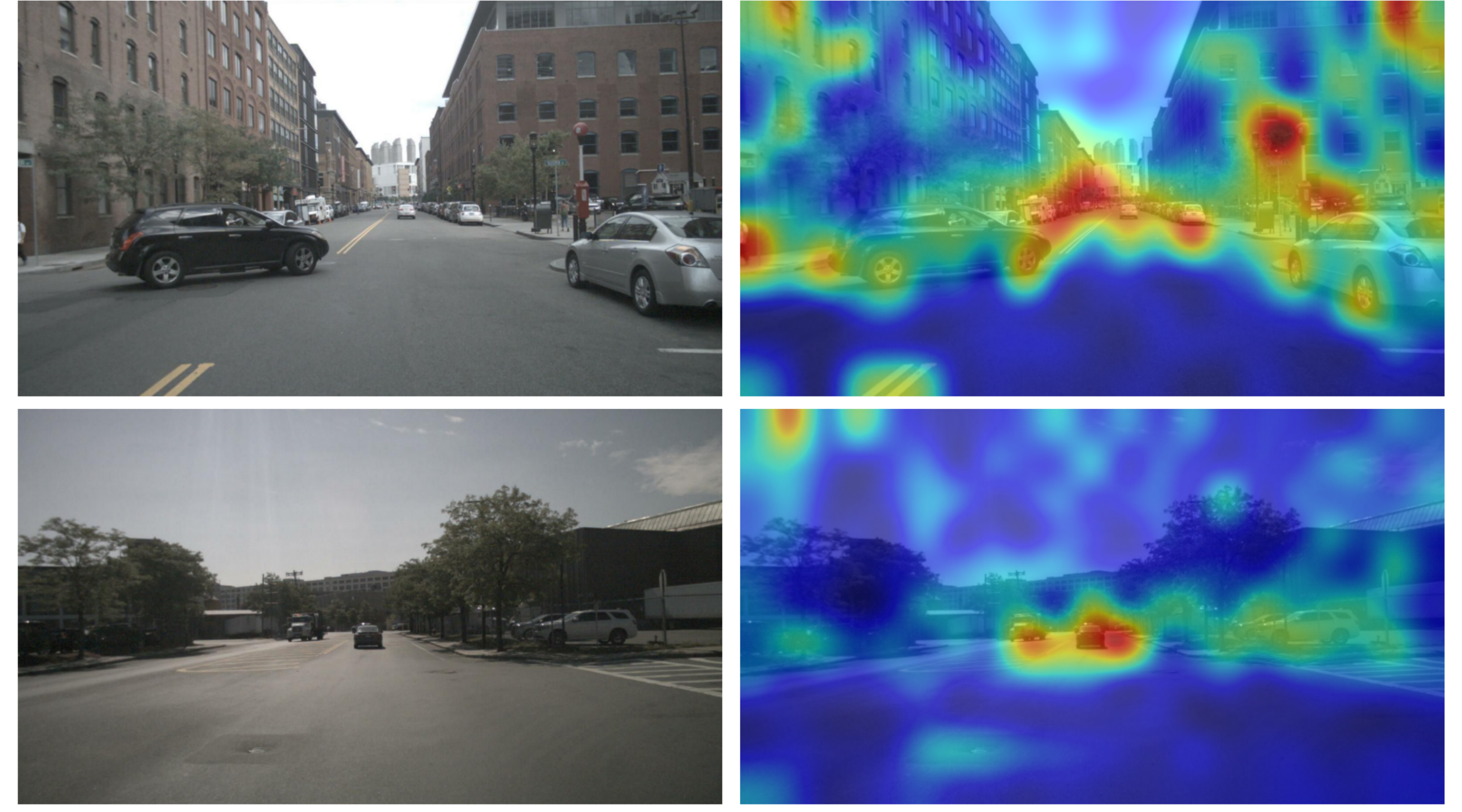}
    \vspace{-3mm}
    \caption{Visualization of the visual anchors.}
    \vspace{-3mm}
    \label{fig:anchor_visual}
\end{wrapfigure}

\subsection{Qualitative Results} 

% \begin{figure}[t]
%     \centering
%     \includegraphics[width=0.99\textwidth]{figures/anchor_visual.pdf}
%     \vspace{-0mm}
%     \caption{Anchor Visual, Focus on driving-related content, such as information about vehicles, Lane marking, Pedestrian and the surrounding roads.}
%     \label{fig:anchor_visual}
%     \vspace{-2mm}
% \end{figure}

Fig.~\ref{fig:visual_compare} presents a qualitative comparison between our method and other baselines on nuScenes. It is evident that our approach significantly outperforms VAD and OmniDrive in long-horizon prediction across diverse weather (sunny/overcast) and road geometries (straight/curved). Specifically, VAD tends to produce short longitudinal predictions, while OmniDrive often exhibits lateral deviations. Additionally, Fig.~\ref{fig:anchor_visual} provides a visualization of the learned visual anchors. The results show that AnchorFormer precisely attends to key elements such as vehicles and roadside pedestrians. This demonstrates the effectiveness of our self-anchored visual distillation in capturing critical environmental features.

\section{Conclusion} 
% We introduce EvoDriveVLA, a novel collaborative perception-planning distillation framework with self- anchored and future-informed distillation for driving. To address the challenges of visual representation degradation and insufficient trajectory precision in existing methods, we propose self-anchored visual distillation to ensure the visual encoder retains its intrinsic perceptual capabilities. Furthermore, we leverage an oracle teacher model integrating privileged future information to provide high-quality trajectory guidance. By incorporating coarse-to-fine iterative refinement and MC-Dropout sampling, the quality of teacher-to-student knowledge transfer is further enhanced. This research establishes a novel paradigm to efficiently distill VLA models for autonomous driving.
We present EvoDriveVLA, a collaborative perception-planning framework featuring self-anchored and future-informed distillation for autonomous driving. To mitigate visual degradation and trajectory imprecision in current models, we employ self-anchored visual distillation to preserve the encoder’s intrinsic perceptual strength. Simultaneously, we leverage an oracle teacher with access to privileged future insights to provide high-fidelity trajectory guidance. Enhanced by coarse-to-fine refinement and MC-Dropout sampling, this approach enables the student to internalize the oracle teacher’s future-predictive capabilities. This research establishes a novel and effective distillation paradigm to enhance VLA models for autonomous driving.

\bibliographystyle{plain}
\bibliography{ref.bib}

\newpage
\appendix

\onecolumn
% \FloatBarrier
\section{Distillation Weight Factor Sensitivity Analysis}
% \FloatBarrier
% \vspace{-30mm}
\begin{table}[H]
  \centering
  \caption{Weight factor sensitivity analysis.}
  \label{tab:Weight Factor}
  \resizebox{0.9\linewidth}{!}{
    \begin{tabular}{l cccc cccc}
    \toprule
    \multirow{2}{*}{Weight Factors}
    & \multicolumn{4}{c}{L2 (m) $\downarrow$} & \multicolumn{4}{c}{Collision (\%) $\downarrow$} \\
    \cmidrule(lr){2-5} \cmidrule(lr){6-9}
    & 1s & 2s & 3s & Avg. & 1s & 2s & 3s & Avg. \\
    \rowcolor{lightgray!50}\multicolumn{9}{c}{\textit{Traj KD Weights(Temp/Weight)}} \\

    6/0.1 & 0.16 & 0.47 & 1.00 & 0.56 & 0.05 & 0.07 & 0.38 & 0.17 \\

    \textbf{5/0.1} & \textbf{0.16} & \textbf{0.45} & \textbf{0.98} & \textbf{0.53} & \textbf{0.02} &\underline{0.05} & \underline{0.38} & \textbf{0.15} \\

    5/0.2 & 0.17 & 0.48 & 1.01 & 0.55& 0.03 & 0.06 & 0.42 & 0.17 \\ 
    
    4/0.1 & 0.16 & 0.46 & 1.00 & 0.54 & 0.03 & 0.03 & 0.38 & 0.15 \\

    3/0.1 & 0.16 & 0.46 & 1.01 & 0.54 & 0.08 & 0.10 & 0.33 & 0.17 \\
    
    \rowcolor{lightgray!50}\multicolumn{9}{c}{\textit{Encoder KD Weights}} \\

    \textbf{0.05} & \textbf{0.16} & \textbf{0.46} & \textbf{1.01} & \textbf{0.54} & \textbf{0.05} & \textbf{0.05} & \textbf{0.38} & \textbf{0.16} \\

    0.1 & 0.17 & 0.49 & 1.04 & 0.57 & 0.07 & 0.07 & 0.39 & 0.18 \\

    0.2 & 0.18 & 0.51 & 1.08 & 0.59 & 0.10 & 0.05 & 0.43 & 0.19 \\

    % 0.2 & \textbf{0.16} & \textbf{0.44} & \textbf{0.96} & \textbf{0.52} & \underline{0.02} & \textbf{0.02} & \underline{0.33} & \underline{0.12} \\

    \bottomrule
    \end{tabular}
  }
  \vspace{0mm}
\end{table}
Tab.~\ref{tab:Weight Factor} provides a weight sensitivity analysis for the distillation of both the trajectory and encoder parts, including the analysis of distillation weights and distillation temperature. Our method maintains stable effects within a reasonable range of weight factors.
% \vspace{-40mm}
\FloatBarrier
\section{Robustness Analysis of the Method under Different Visual Conditions}
\begin{table}[H]
  \centering
  \caption{Verification of perceptual robustness}
  \label{tab:Robustness}
  \resizebox{0.9\linewidth}{!}{
    \begin{tabular}{l cccc cccc}
    \toprule
    \multirow{2}{*}{Methods} 
    & \multicolumn{4}{c}{ST-P3 L2 (m) $\downarrow$} & \multicolumn{4}{c}{UniAD L2 (m) $\downarrow$} \\
    \cmidrule(lr){2-5} \cmidrule(lr){6-9}
    & 1s & 2s & 3s & Avg. & 1s & 2s & 3s & Avg. \\
    \cmidrule(lr){1-9}

    Rainy/Night & 0.12 & 0.23 & 0.38 & 0.24 & 0.16 & 0.41 & 0.76 & 0.43 \\

    Sunny & 0.15 & 0.24 & 0.40 & 0.27 & 0.17 & 0.36 & 0.80 & 0.44 \\ 
    
    Rainy/Night+KD & 0.10 & 0.21 & 0.35 & 0.22 & 0.13 & 0.39 & 0.68 & 0.40 \\

    Sunny+KD & 0.15 & 0.22 & 0.37 & 0.25 & 0.17 & 0.36 & 0.78 & 0.44 \\
    
    \bottomrule
    \end{tabular}
  }
  \vspace{0mm}
\end{table}
Tab.~\ref{tab:Robustness} demonstrates the visual robustness of our method, where visual distillation significantly improves prediction accuracy. Experiments conducted under challenging conditions (such as rainy, cloudy, and night environments) show that our method maintains stable performance, confirming our claims about perceptual robustness across different scenes.
\section{Limitations}
Our evaluation is primarily conducted on the nuScenes and NAVSIM benchmarks, which predominantly feature driving scenes from specific geographic regions. Further validation on more diverse datasets with varying weather conditions and traffic rules would help to verify the model's zero-shot generalization across different countries.
\section{Impact Statements} 
This research focuses on the technical advancement of knowledge distillation for autonomous driving models and is conducted using publicly available, anonymized datasets. As a fundamental study in algorithmic framework design, it does not present any direct negative societal impacts or immediate ethical concerns.
% \FloatBarrier

% \FloatBarrier
\newpage
\section{Dataset Examples}
% \vspace{-100mm}
\begin{figure}[H]
    \centering
    \includegraphics[width=1.05\textwidth]{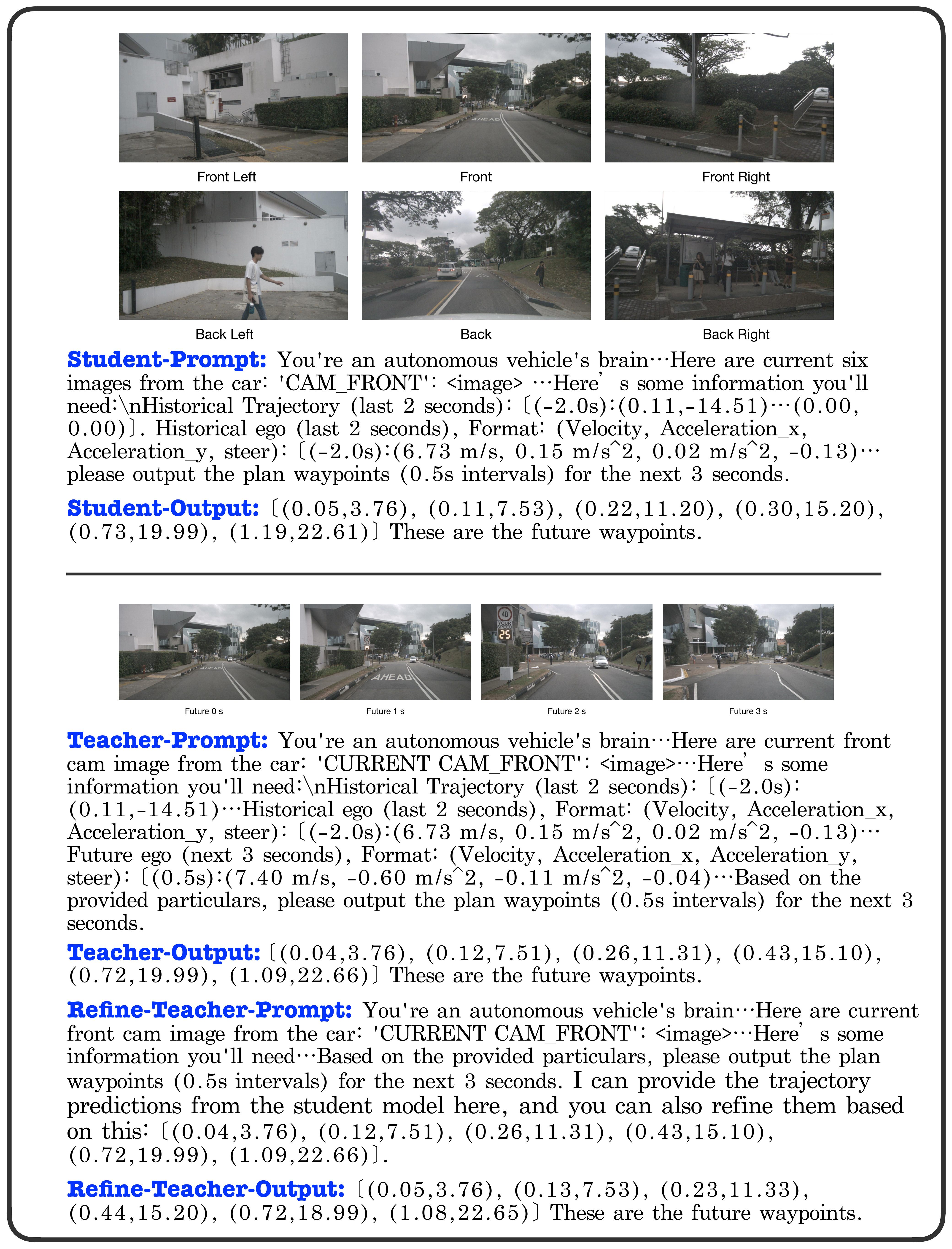}
    \vspace{15mm}
    % \caption{Student model input images, current six views.}
    \label{fig:sample0}
    \vspace{-0mm}
\end{figure}
% % \subsection{50f90acff4cd4ac5bdd62f8bd2728878}
\begin{figure}[H]
    \centering
    \includegraphics[width=1.05\textwidth]{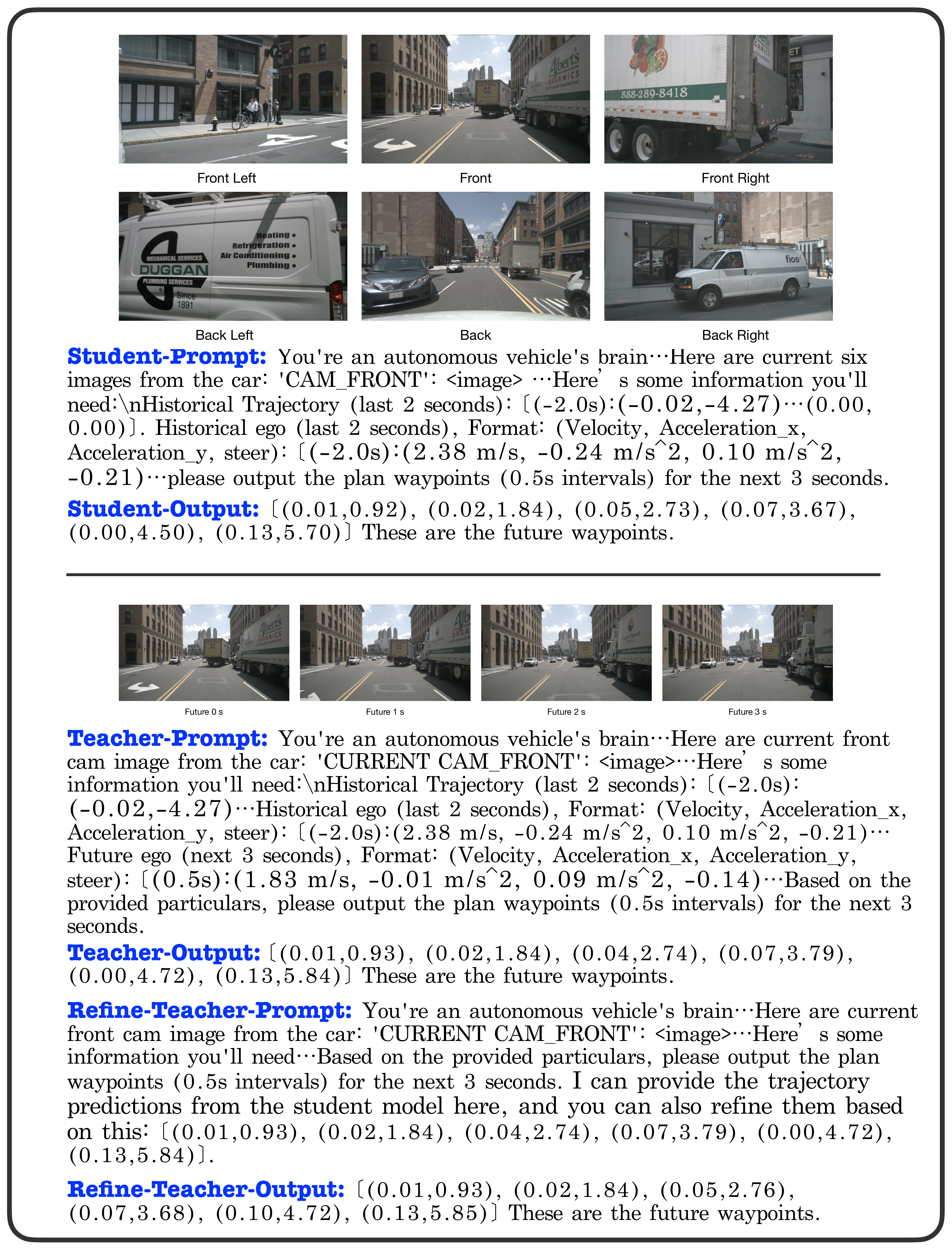}
    \vspace{-20mm}
    % \caption{Student model input images, current six views.}
    \label{fig:sample1}
    \vspace{-0mm}
\end{figure}
% % \subsection{5745a861cde248e3b6fb6d6a68da8309}
\begin{figure}[H]
    \centering
    \includegraphics[width=1.05\textwidth]{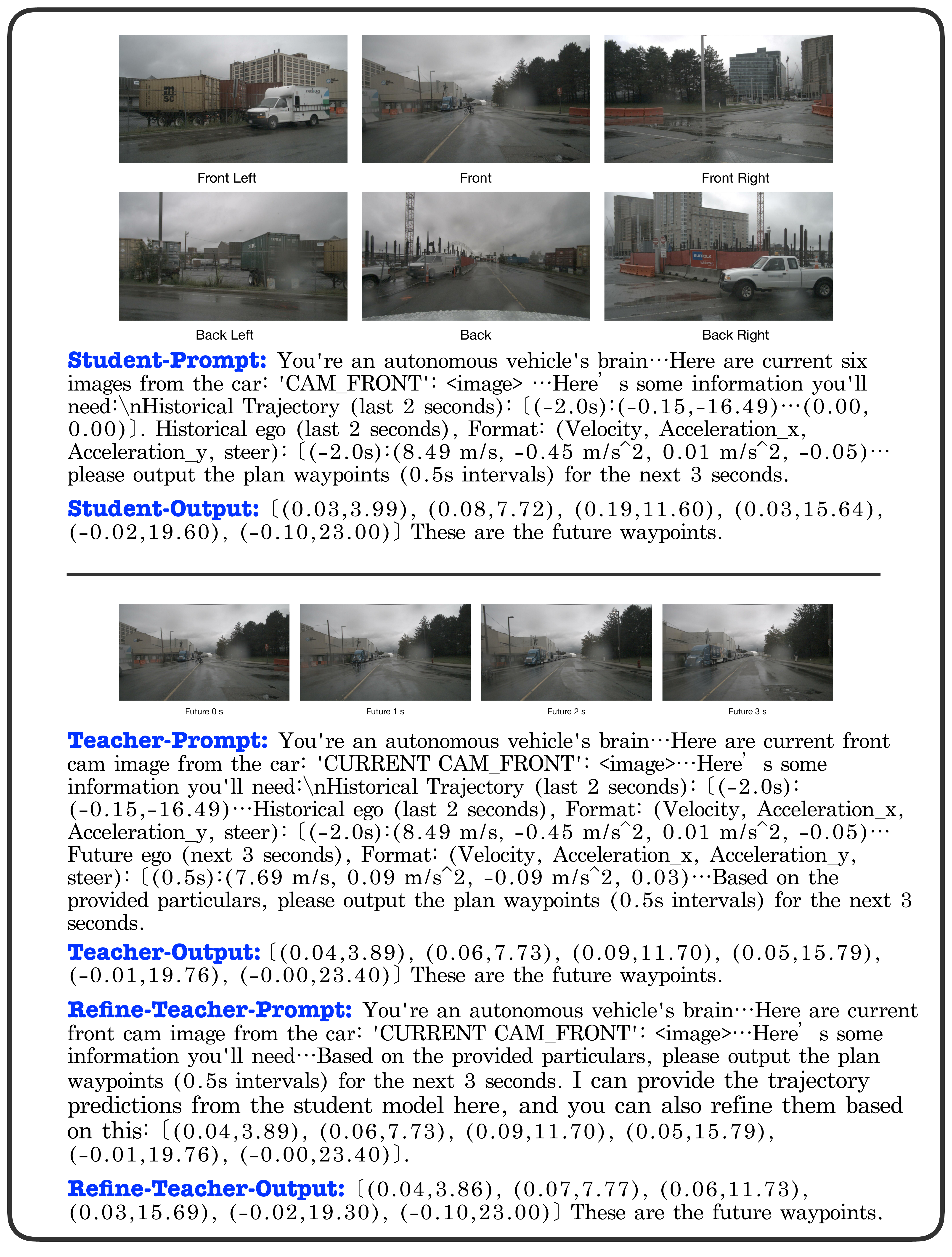}
    \vspace{-20mm}
    % \caption{Student model input images, current six views.}
    \label{fig:sample2}
    \vspace{-0mm}
\end{figure}

\section{Additional Qualitative Results}
% \vspace{-40}
\begin{figure}[H]
    \centering
    \includegraphics[width=1.05\textwidth]{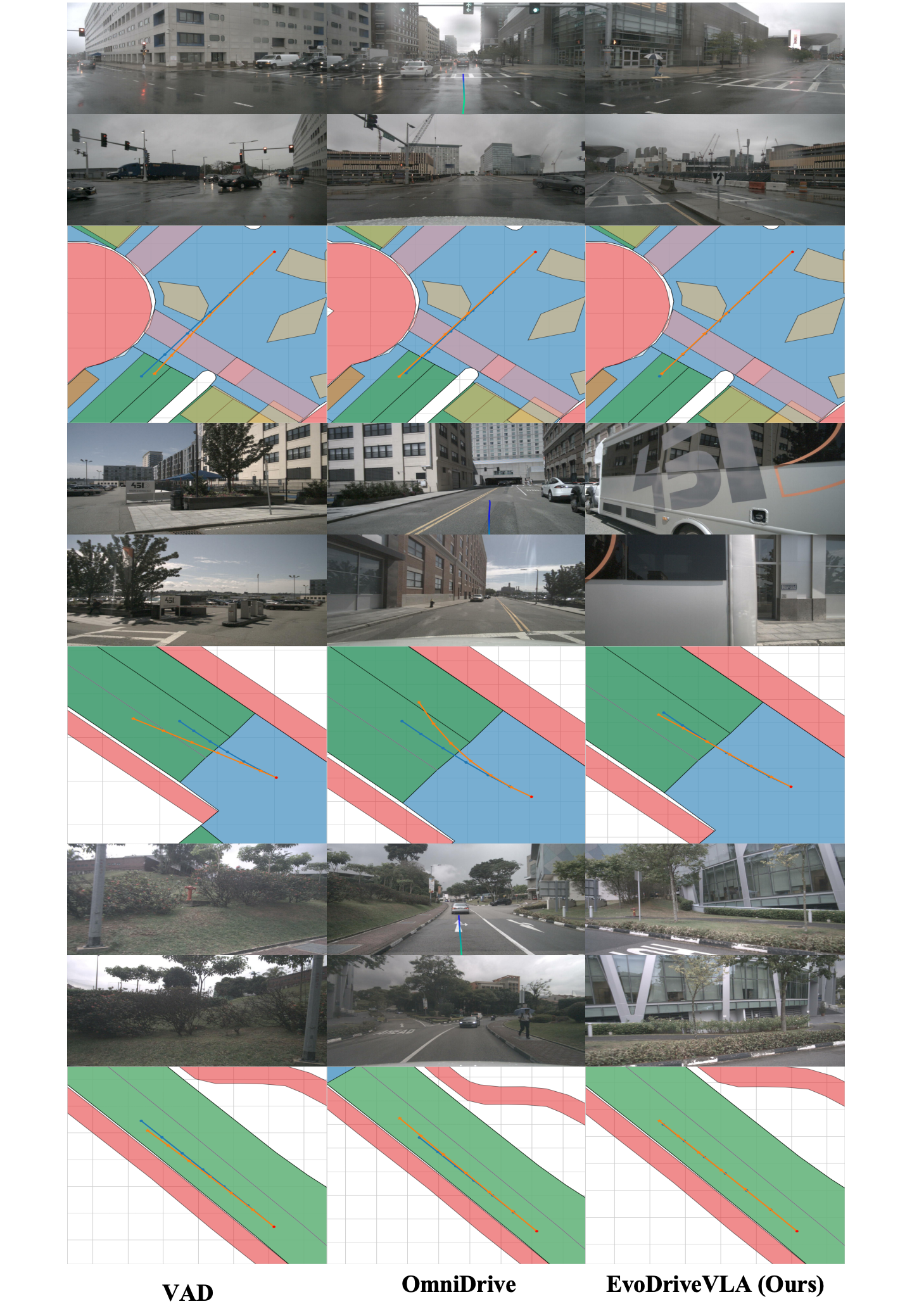}
    \vspace{-5mm}
    % \caption{Student model input images, current six views.}
    \label{fig:visual_APP_01}
    \vspace{-0mm}
\end{figure}

\begin{figure}[H]
    \centering
    \includegraphics[width=1.05\textwidth]{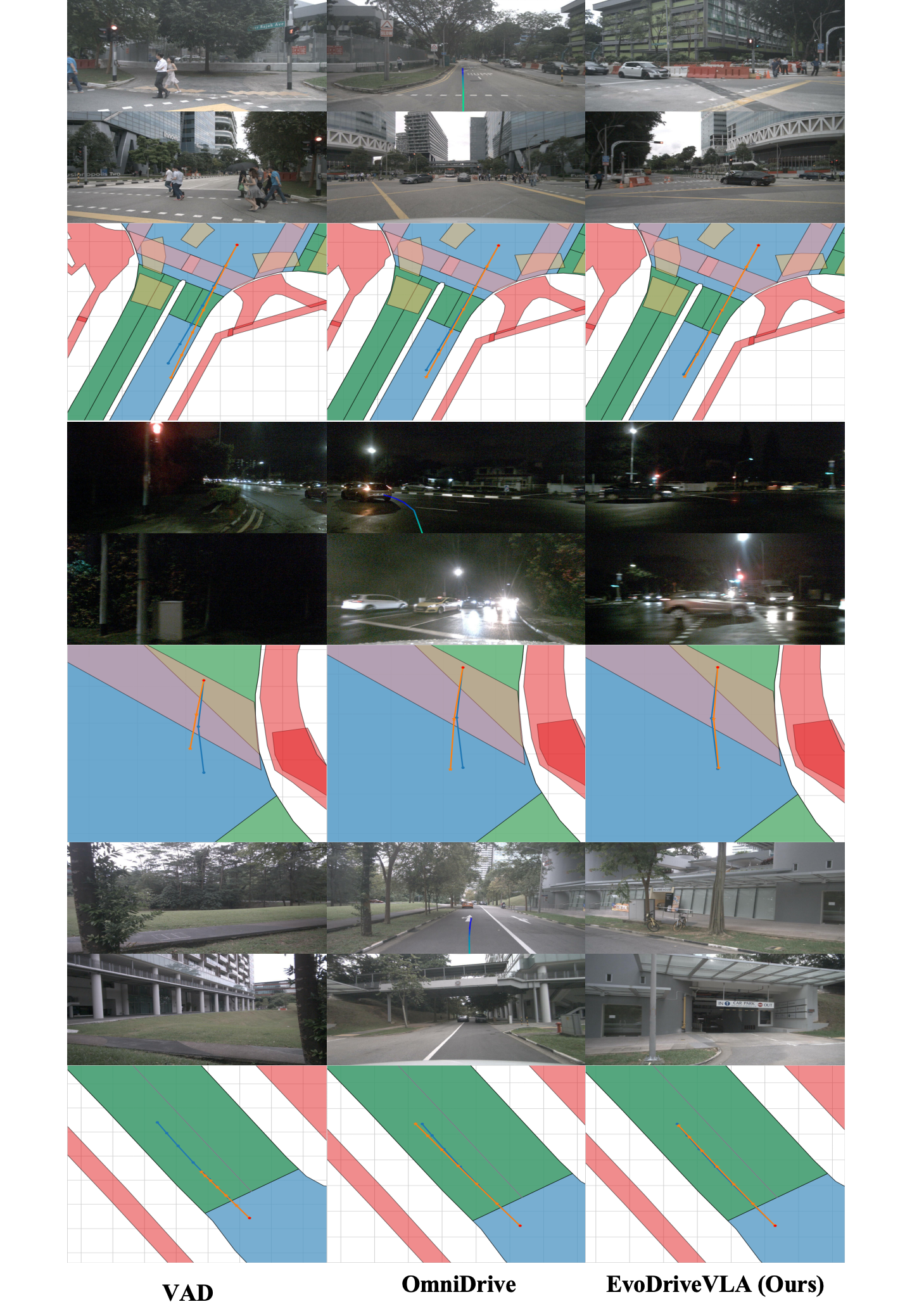}
    \vspace{-0mm}
    % \caption{Student model input images, current six views.}
    \label{fig:visual_APP_02}
    \vspace{-0mm}
\end{figure}

% \section{Technical appendices and supplementary material}
% Technical appendices with additional results, figures, graphs, and proofs may be submitted with the paper submission before the full submission deadline (see above). You can upload a ZIP file for videos or code, but do not upload a separate PDF file for the appendix. There is no page limit for the technical appendices. 

% Note: Think of the appendix as ``optional reading'' for reviewers. The paper must be able to stand alone without the appendix; for example, adding critical experiments that support the main claims to an appendix is inappropriate. 

% \input{tables/Com_FSDrive}

%%%%%%%%%%%%%%%%%%%%%%%%%%%%%%%%%%%%%%%%%%%%%%%%%%%%%%%%%%%%

% \newpage
% \input{checklist.tex}

\end{document}